\pdfoutput=1

\documentclass[10pt, a4paper]{article}

\usepackage[final]{lrec2026}

\usepackage{latexsym}

\usepackage{paralist} 

\usepackage[T1]{fontenc}

\usepackage[utf8]{inputenc}

\usepackage{microtype}

\usepackage{inconsolata}

\usepackage{graphicx}

\title{From \textit{\rm \it Trial by Fire} To \textit{\rm \it Sleep Like a Baby}: A Lexicon of Anxiety Associations for 20k English Multiword Expressions}

\name{Saif M. Mohammad}

\address{National Research Council Canada\\
  \texttt{saif.mohammad@nrc-cnrc.gc.ca} \\}

\newcommand{\bl}[1]{{\color{black} #1}}

\abstract{
Anxiety is the unease about a possible future negative outcome. In recent years, there has been growing interest in understanding how anxiety
relates to our health, well-being, body, mind, and behaviour. 
This includes work on lexical resources for word--anxiety association.
However, there is very little anxiety-related work on larger units of text such as multiword expressions (MWE).
Here, we introduce the first large-scale lexicon capturing descriptive norms of anxiety associations for more than 20k English MWEs. 
We show that the anxiety associations are highly reliable.
We use the lexicon to study prevalence of different types of anxiety- and calmness-associated MWEs; and how that varies across two-, three-, and four-word sequences. We also study the extent to which the anxiety association of MWEs is compositional (due to its constituent words).
The lexicon enables a wide variety of anxiety-related research 
in psychology, NLP, public health, and social sciences. The lexicon is freely available: \url{https://saifmohammad.com/worrylex.html}\\
\newline   }
\begin{document}
\maketitleabstract


\section{Introduction}

Anxiety is the feeling of unease or nervousness one feels because they anticipate some possible future negative outcome. 
It is a normal part of the human experience and has evolved for our benefit:
the sense of unease can be a useful driver of actions that lead to safety, security, prosperity, and social harmony \cite{bateson2011anxiety}.
Further, anxiety (both normal and clinical) and, its converse, calmness have been been shown to impact productivity, cognitive outcomes (e.g., accuracy on mental tasks), and even our physical bodies \cite{kim2005psychobiology,siddaway2018reconceptualizing}.
Thus, anxiety and calmness 
are active areas of research in Psychology, Medicine, Public Health, Social Science, Economics, and even the Humanities. 

Language is a powerful medium for expressing anxiety (consciously and unconsciously) and language-resource work has produced large repositories of word--emotion associations and sentences annotated for emotions (including for anxiety). However, we are not aware of any large-scale work on multiword expressions (MWEs) and anxiety. MWEs have been defined with some differences in past works, but here we simply consider sequences of two or more words (often with some interesting semantic, syntactic, or functional property) as MWEs. 
Broadly speaking, MWEs are important in NLP, linguistics, social sciences, and psychology because meaning is often not compositional \cite{smolka2020role} and MWEs reveal insights about the structure of language, social interaction, and cognitive processing. Yet, unlike their lexical or sentence cousins, far fewer language resources exist for MWEs.
Our work at the intersection of anxiety and MWEs makes three contributions: \\[-8pt]
\setdefaultleftmargin{1em}{}{}{}{.5em}{.5em}
\begin{compactenum}
    \item We make a comprehensive case for why anxiety- and calmness-associations of MWEs matters in various disciplines (\S 3).\\[-8pt]
    \item We conduct a large crowdsourcing effort to determine anxiety- and calmness associations for more than 20k English MWEs (\S 4). We will refer to it as the WorryMWEs lexicon. The lexicon captures descriptive norms (how people use language and how they perceive MWE--anxiety associations); these are not normative (how people should use language).\\[-8pt]
    \item We use WorryMWEs to study prevalence of different types of anxiety- and calmness-associated MWEs (\S 5.1). We also study the extent to which the anxiety association of MWEs is compositional (due to its constituent words) (\S 5.2).\\[-8pt]
\end{compactenum}

\noindent 
WorryMWEs (with entries for 20k MWEs) is bundled together with WorryWords (with entries for 44k words) and released as WorryLex v2.\footnote{https://saifmohammad.com/worrylex.html}

\section{Related Work}

MWE work in NLP has largely focused on the automatic discovery, processing, and understanding of MWEs from corpora (see surveys such as \cite{constant-etal-2017-survey,smolka2020role}). Relatively less work has gone into manually compiling large lists of MWEs. Notable in such efforts is work by 
\newcite{PMID:35867207}, who manually compiled a lexicon of about 62,000 MWEs. 
They added concreteness ratings for the MWEs, as well as the frequencies of the MWEs in a subtitles corpus \cite{brysbaert2012adding}.
\newcite{takahashi2024comprehensive} compiled a lexicon of 160,000 Japanese MWEs.
\newcite{tong2024metaphor} compiled a 10k English metaphor--literal paraphrase pairs dataset.
There is even less work on annotating MWEs for information such as sentiment, despite work on many *word*--emotion association lexicons for a handful of categorical emotions (as in the NRC Emotion Lexicon \cite{MohammadT10,MohammadT13}) and for valence, arousal, and dominance (\citet{warriner2013norms} and \citet{mohammad-2018-obtaining} for English;  \citet{moors2013norms} for Dutch, and  \citet{vo2009berlin} for German). 
The recently released NRC VAD Lexicon v2 includes valence, arousal, and dominance entries for 10k MWEs and 44k words \cite{mohammad2025breaking}.
\newcite{jochim-etal-2018-slide} crowdsourced the sentiment annotation of 5000 English idioms.
\newcite{ibrahim2015idioms} annotated 3600 Modern Standard Arabic idioms for sentiment. 

\newcite{worrywords-emnlp2024} created a word--anxiety association lexicon for 44,000 English words.\footnote{https://saifmohammad.com/worrywords.html} The lexicon has been used to detect and study human values \cite{yeste2026schwartz}, temporal trends in the expression of anxiety on social media \cite{mohammad2026temporalanxiety}, and mental health memes \cite{mazhar2025memes}.

Going beyond words, there exist some sentence/post-level datasets of social media \textit{posts} manually annotated with anxiety or stress labels \cite{rastogi2022stress,mitrovic-etal-2024-comparing,turcan-mckeown-2019-dreaddit}.
However, there are no large datasets of MWEs annotated for anxiety. 
We discuss some challenges in creating such lexicons and how such lexicons can be widely useful, in the next section.




\section{Why Anxiety- and Calmness- Associated MWEs Matter}

Even though there is considerable NLP work on words (such as the creation of large lexicons and lexical-semantic analysis) and on sentences (such as the
creation of large corpora, sentences annotated for emotions, and automatic systems for sentence classification), there is relatively much less work on multiword expressions. 
Arguably, this is because it is easier to deal with words and sub-word units in NLP, whereas
MWEs blur the boundary between vocabulary and grammar by often acting as a single unit and 
having varying levels of fixedness.
(Also see \textit{Multiword expressions: A pain in the neck for NLP} \cite{sag2002multiword}.)
Yet, MWEs are important in NLP, linguistics, social sciences, and psychology because their meaning is often not compositional and so they can reveal insights about the structure of language, social interaction, and cognitive processing. 
See \newcite{ramisch2015multiword,sag2002multiword,smolka2020role} for a discussion on the applications of MWE resources in general.
Below, we make a case for why research on anxiety-related MWEs matters and how anxiety lexicons can be useful (even in the era of large language models).\footnote{Note that all of the work described for the disciplines below are relevant to NLP, and so we do not list a separate bullet point just for NLP.}

\begin{compactenum}
    
    \item \textit{Semantics and Human--Computer Interaction (HCI).}
MWEs such as \textit{on edge}, \textit{at peace with myself, jumping out of my skin}, and \textit{keep your cool}  have non-literal meanings that cannot be determined simply from the meanings of their constituents.
They can be especially revealing in how people store and retrieve language chunks of meaning.
Thus, both linguists and developers of artificial chat agents benefit from a large repository of MWE--anxiety associations. 

\item \textit{Pragmatics, Emotions, and Framing.}
MWEs are often laden with rich connotative meaning, conveying subtle nuances of emotional intensity, politeness, formality, or social acceptability \cite{sag2002multiword,zgusta1967multiword,citron2019idiomatic,allawama2025idioms}. MWEs such as \textit{losing control, on edge,} \textit{can’t take it anymore} (anxiety), \textit{at peace with myself}, or \textit{taking it in stride} (calmness) encode folk psychological concepts of emotional states.
These expressions reveal how people naturally talk about and categorize emotion, which helps researchers build models of affect.
MWEs often have an emotional punch, thereby influencing perception, recall, and judgment \cite{citron2019idiomatic}.
Thus, MWEs like \textit{war on drugs} or \textit{family values} are used to frame complex issues in persuasive ways. Therefore, MWEs, especially those associated with anxiety, are highly relevant to analyzing online discourse and political rhetoric.

\item \textit{Cognitive Science, Cognitive Linguistics, Cognitive Psychology.}
MWEs often draw on physical and embodied metaphors \cite{kacinik2014sticking}. Examples of anxiety and calmness-related embodied MWEs include: 
\textit{jumping out of my skin}, \textit{butterflies in my stomach}, \textit{my heart’s racing},
\textit{smooth sailing}, \textit{cool as a cucumber}, and \textit{a weight lifted}. 
Thus MWEs are a window into embodied and metaphorical thinking. 
Anxiousness associated MWEs can be used to study how emotional language is grounded in sensorimotor experience and is organized metaphorically.

\item \textit{Lexicon and Mental Representation.}
Many emotionally charged MWEs are stored as chunks in the mental lexicon \cite{villavicencio2004lexical}.
Their processing shows that language users treat them as formulaic units, not constructed from scratch each time.
Anxiousness associated MWEs can be used to study how emotional language is lexically represented, and whether they facilitate faster retrieval and use in communication (thereby providing evolutionary benefits).

\item \textit{Affective Science.}
MWEs frequently reflect emotion regulation strategies, both maladaptive  and adaptive \cite{nichter2010idioms,lee2017figurative,cole2010role}. Examples of anxiety-associated MWEs pertaining to emotion regulation, include: \textit{bottling it up, trying to push it down, taking a deep breath,} and \textit{letting it go}.
They give insight into implicit self-regulatory processes people engage in during emotional episodes.
MWEs provide linguistic evidence for how affect interacts with attention, memory, appraisal, and prediction—core components of emotion theories.
Anxiety MWEs often reflect future-oriented, uncertain, or threat-focused cognition (e.g., \textit{what if something goes wrong?}).
Calmness MWEs often reflect present-centered, grounded cognition (e.g., \textit{taking things as they come}).
WorryMWEs can be used to study emotion regulation strategies and inform theories of emotion by showing how people encode appraisals and attention patterns in language.

\item \textit{Computational Social Science (CSS).} Anxiety and fear are particularly relevant to CSS in recent years with growing anxiety over issues such as climate change, immigration, health care, and civil rights. On the one hand, a healthy level of anxiety can push people towards collective action to solve big problems; yet, on the other hand, heightened anxiety can be used as a manipulative tool by politicians for votes, and to take away rights and civil liberties \cite{cap2016language}.
Large MWE--anxiety lexicons can help track levels of anxiety towards these issues in various stake holders.  

\item \textit{Sociolinguistics and Discourse Analysis.}
Expressions and norms of anxiousness vary across cultures, communities, and situational contexts \cite{heinrichs2006cultural}. This applies to MWEs as well: for example, in North America one may commonly hear \textit{spinning out} and \textit{losing it},
whereas in India one may come across \textit{my head is heavy} (overwhelming emotions).
Thus MWEs associated with anxiousness and calmness can act as discursive markers of identity, emotional stance, and cultural framing of emotional experience.
Anxiousness MWEs local to different communities can be used to study social norms of emotional regulation or cultural values around anxiousness and composure.

\item \textit{Corpus Linguistics and Public Health Research.}
Linguists and public health researchers can analyze the frequency and context of MWEs linked to emotional states in clinical narratives, social media, and everyday speech.

\item \textit{Sentiment and Emotion Analysis.}
Availability of an MWE--anxiety lexicon can improve sentiment analysis, emotion detection, and affective computing models.

\item \textit{Construction Grammar and Language Acquisition.}
Some MWEs linked to anxiety or calmness occur in fixed syntactic patterns: e.g.,
\textit{I can’t take it anymore}, \textit{I'm losing control}, \textit{I finally feel at ease}, and \textit{It’s going to be okay}.
Both children and second-language learners rely heavily on formulaic expressions for fluent communication \cite{jackendoff1997twistin}. Thus, anxiety-related MWEs can be used to study 
 how some emotional expressions become fixed over time as well as how
children acquire and develop the language of emotions.
\end{compactenum}
\noindent Thus language resources at the intersection of MWEs and anxiety are foundational to our understanding of a wide variety of phenomena. Note that automatic prediction of anxiety from individual text instances is only a small part of the use cases. In addition to presenting the broad overview above, in Sections 5.1 and 5.2 ahead, we show how we use WorryMWEs to obtain new insights about MWEs (insights that are most directly answered by the lexicon rather than by using some ML system or LLMs). Finally, even though LLMs can be prompted to generate anxiety lexicons, any inferences drawn from an automatic approach requires manual validation. Portions of the manually created anxiety lexicon presented here can be used to improve the generations of the LLM and held out portions can be used to validate the LLM generations. 

\section{MWE Annotations for Anxiety}

Key steps in creating the MWE--anxiety lexicon: 
\begin{compactenum}
\item Selecting the terms to be annotated 
\item Developing the questionnaire
\item Developing measures for quality control (QC)
\item Annotating terms on a crowdsource platform
\item Discarding data from outlier annotators (QC)
\item Aggregating data from multiple annotators to determine the anxiety association scores
\item Determining reliability (quality) of annotations
\item Distribution
\end{compactenum}
We describe each of these steps below.\\[3pt]
\noindent {\bf 1. Term Selection.}
\noindent \bl{
We wanted to include various kinds of MWEs, including common phrases, light verb constructions, and idiomatic constructions. 
However, identifying MWEs from a large corpus of text is not trivial.
Further, we wanted to include terms for which other linguistically interesting annotations already exist (such as concreteness ratings). 
Thus, for our anxiety associations work we chose the 10,000 most frequent MWEs compiled by \newcite{PMID:35867207}.} These were all bigrams (two-word sequences) and were annotated first as described in the steps below. We then selected an additional 10,600 most frequent trigrams (three-word sequences) and fourgrams (four-word sequences) for annotation.\\[-7pt]

\noindent{\bf 2. Anxiety Questionnaire.}
The questionnaire used to annotate the data 
 was essentially the same as that used for the WorryWords project \cite{worrywords-emnlp2024}, with minor changes such as replacing ``words" with ``terms" or "MWEs", and including MWEs in the set of examples provided. 
 Detailed directions
 and example questions (with suitable responses) were provided. 
 The primary instruction and the question presented to annotators is shown below.\\[-12pt]

{
\noindent\makebox[\linewidth]{\rule{0.5\textwidth}{0.4pt}}\\
{ \small
\noindent Consider anxiety to be a broad category that includes:\\
\indent \textit{jittery, antsy, insecure, nervous, unease, unnerving,\\ 
\indent  worried, tense, nerve-racking, apprehensive, troubled,\\
\indent fretful, self-doubting, discontented, concerned}\\ 
Consider calmness to be a broad category that includes:\\
\indent \textit{calm, relaxed, comforted, serene, at ease, peaceful,\\ 
\indent carefree, composed, collected, untroubled, contented,\\
\indent self-assured, unconcerned, indifferent, uninvolved}\\[4pt]
 \noindent  If you do not know the meaning of a multiword expression or are unsure, you can look it up in a dictionary (e.g., the Merriam Webster) or on the internet.\\
\noindent  \textbf{Quality Control:} Some questions have pre-determined correct answers. If you mark these questions incorrectly, we will give you immediate feedback in a pop-up box. An occasional misanswer is okay. However, if the rate of misanswering is high (e.g., >20\%), then all of one's HITs may be rejected.

\noindent Select the options that most English speakers will agree.\\[4pt]
\noindent \textbf{Q1.  <term> is often associated with feeling:}\\[-1pt]
\indent 3: very anxious \hspace{14mm} -1: slightly calm\\[-1pt]
\indent 2: moderately anxious \hspace{5mm}  -2: moderately calm\\[-1pt]
\indent 1: slightly anxious \hspace{10mm}  -3: very calm\\[-1pt]
\indent 0: not associated with feeling anxious or calm\\[-6pt]
}
\noindent\makebox[\linewidth]{\rule{0.5\textwidth}{0.4pt}}\\[-7pt]

}

\noindent{\bf 3. Quality Control Measures.}
We annotated 2\% of the data ourselves first and interspersed these with the rest. We refer to these questions as \textit{gold} (aka \textit{control}) questions. 
Half of the gold questions were used to provide immediate feedback to the annotator (in the form of a pop-up on the screen) in case they mark them incorrectly. We refer to these as \textit{popup gold}. This helps prevent the situation where one annotates a large number of instances without realizing that they are doing so incorrectly. 
As an annotator may unbeknownst to us share answers to gold questions with others (despite this being against the terms of annotation), 
the other half of the gold questions were also separately used to track how well an annotator was doing the task. For these gold questions no popup was displayed in case of errors. 
We refer to these as 
\textit{no-popup gold}.\\[-7pt]

\noindent{\bf 4. Crowdsourcing.} 
We setup the anxiety annotation task on the crowdsourcing platform, {\it Mechanical Turk}.
 In the task settings, we specified that we needed annotations from nine people for each MWE.
We obtained annotations from native speakers of English residing around the world. Annotators were free to provide responses to as many terms as they wished. 
The purpose of the task and how their annotations will be used was made clear, and consent was obtained.
The annotation task was approved by 
our institution's review board.\\[-7pt]

\noindent {\it Demographics:} About 83\% of the respondents who annotated the MWEs live in USA. The rest were from India, United Kingdom, and Canada. 
The average age of the respondents was 35 years. Among those that disclosed their gender, about 58\% were female and about 42\% were male.\footnote{Respondents were shown optional text boxes to disclose their demographic information as they choose; especially important for social constructs such as gender, in order to give agency to the respondents and to avoid binary language.} 

\noindent{\bf 5. Filtering.} 
If an annotator's accuracy on the gold questions (popup or non-popup) fell below 80\%, then all of their annotations were discarded.\\[-7pt] 

\noindent{\bf 6. Aggregation.} 
Every response was mapped to an integer from -3 (very calm) to 3 (very anxious). 
The final anxiety score for each term is the average score it received from each of the annotators.
We also created a categorical version of the lexicon by labeling MWEs that got a score $\geq 2.5$ as associated with \textit{high anxiety}, $\geq 1.5$ and $< 2.5$ as \textit{moderate anxiety}, $\geq 0.5$ and $< 1.5$ as \textit{slight anxiety},
$> -0.5$ and $< 0.5$ as \textit{neither anxiety nor calmness}, and so on.  
We refer to the
list of MWEs
along with their 
 real-valued 
final scores and categorical labels as the {\it WorryMWEs Lexicon}. 
 See Table~\ref{tab:ann} for summary statistics. We include details from the WorryWords dataset (past work on English unigrams by \newcite{worrywords-emnlp2024}) for easy reference.

\begin{figure*}[t]
	     \centering
	     \includegraphics[width=\textwidth]{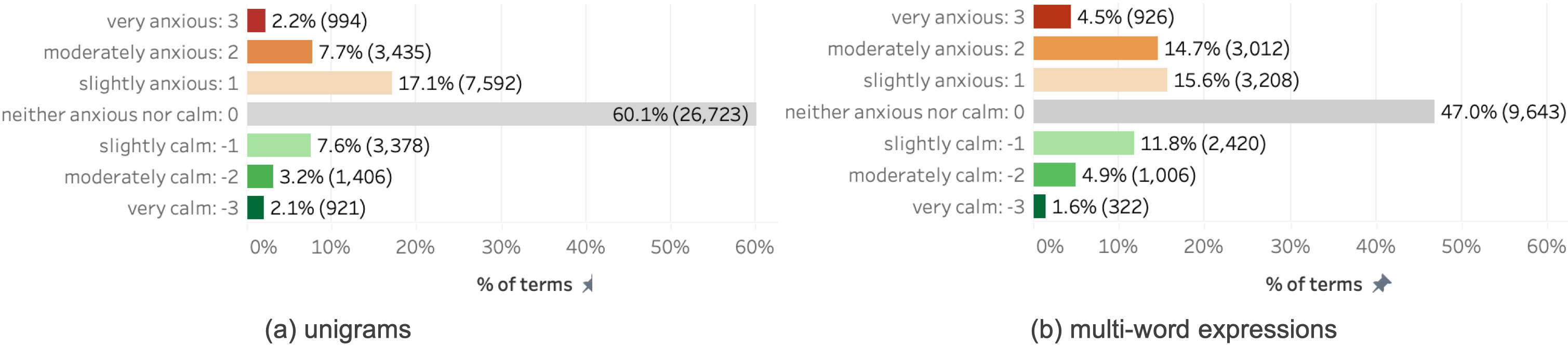}
	   \vspace*{-5mm}
      \caption{Distribution of terms in (a) WorryWords (unigrams) and (b) WorryMWEs (MWEs): percentage of terms associated with each class. (The total number is shown in parenthesis.)}
	     \label{fig:WorryWords-distrib}
	 \end{figure*}

Figure \ref{fig:WorryWords-distrib} (a) shows the distribution of the different classes \bl{for the $\sim$44k words} in the WorryWords lexicon;
whereas (b) shows the distribution of the different classes \bl{for the $\sim$20k words} in the WorryMWEs lexicon we created. We observe that
MWEs are markedly more polar (associated with some degree of anxiety or calmness): 53\% of the MWEs; compared to unigrams: 39.9\% of the unigrams. 
Further, almost twice as many MWEs are associated with anxiousness ($\sim$36\%) than with calmness ($\sim$18\%).
These numbers are consistent with the idea that MWEs are an important linguistic mechanism to convey anxiety and calmness.\\[-7pt]

\begin{table*}[t!]
\caption{\label{tab:ann} {A summary of the WorryWords and WorryMWEs annotations. MAT = mean  annotations per term. SHR, measured through both Spearman rank and Pearson's correlations (last two columns), indicate high reliability.}
\vspace*{2mm}
}
\centering
\small{
\begin{tabular}{llrrrrrrrrr}
\hline 

{\bf Dataset} 	&\bf Terms & \bf \#Terms	   & \bf \#Annotators &\bf \#Annotations &\bf MAT  &\bf SHR ($\rho$)		 &\bf SHR ($r$) \\\hline
WorryWords &unigrams 		& 44,450   & 1,020 & 375,796  & 8.45 	& 0.82 &0.89\\
WorryMWEs &bigrams 		& 9,891 & 217 &72,286 	&7.31 &0.81 &0.89\\
          &trigrams 	& 7,287  &285  &53,174 	&7.30 &0.93 &0.96\\
        &fourgrams 		& 3,352 &229  &24,301 	&7.25 &0.95 &0.96\\
 \hline
\end{tabular}
}
 \vspace*{2mm}
\end{table*}

\noindent \textbf{7. Assessing Reliability of the Annotations} 
We measure quality of the produced lexicon through a metric of reproducibility---repeated independent manual
annotations from multiple respondents should result in similar scores.
To assess this reproducibility, we calculate
average {\it split-half reliability (SHR)} over 1000 trials. Note that SHR is a common way to determine reliability of responses used to generate scores on an ordinal scale 
\cite{weir2005quantifying}. 
All annotations for an
item are randomly split into two halves. Two separate sets of scores are aggregated, just as described in bullet 6 above, from the two halves. 
Then we determine how close the two sets of scores are (using a metric of correlation). 
This is repeated 1000 times and the correlations are averaged.
The last two columns in Table~\ref{tab:ann} show the results (split half-reliabilities). Scores for WorryWords are shown for comparison. SHR scores of \bl{0.81} (Spearman rank correlation) and 0.89 (Pearson correlation) for bigrams indicate high reliability of the aggregated real-valued anxiety values obtained from the annotations.\footnote{For reference, if the annotations were random, then repeat annotations would lead to an SHR of 0. Perfectly consistent repeated annotations lead to an SHR of 1.} These scores are comparable to those reported in \newcite{mohammad-2024-worrywords} for WorryWords.
The scores for trigrams and fourgrams are even higher (greater than 0.9).\\[-7pt]

\noindent \textbf{8. Distribution.} WorryMWEs is freely available for research. Terms of use are listed on the project page.\footnote{\url{https://saifmohammad.com/worrylex.html}}
WorryMWEs (with entries for 20k MWEs) is bundled together with WorryWords (with entries for 44k words) and released as WorryLex v2 \cite{WorryLex}.\footnote{v2.0 included about 10k MWEs; whereas v2.1 includes more than 20k MWEs.}

\section{Using WorryMWEs and WorryWords to Study Anxiety}

\bl{Even though anxiety is a common emotion, there is much that we do not know in terms of how it manifests in language (in words, in expressions, in grammatical function, etc.). 
As part fo this work, we use WorryMWEs and WorryWords to explore:
\begin{compactenum}
    \item What is the distribution of anxiety and calmness classes in different types of words and MWEs?
    \item To what extent do neutral and anxious/calmness terms come together to create anxious/calmness MWEs (anxiousness compositionality)?
\end{compactenum}
}
\noindent We explore these in the subsections below.


\subsection{How Common are the Anxiousness and Calmness Classes in Different Types of Words and MWEs?}

A broad question of interest at the intersection of language and anxiety, is how language conveys anxiety. 
We use WorryWords to explore the extent to which we see anxiety- and calmness-associated words in different categories of grammatical function in English; and
WorryMWEs to explore the extent to which we see anxiety- and calmness-associated words in different types of MWEs. 

\noindent \textbf{Types of Words:} We determined grammatical category, specifically the most common parts of speech (PoS) associated with a word, from the SUBTLEX corpus \cite{brysbaert2009moving} (SUBTitled movie and TV corpus). SUBTLEX has word frequency and PoS information for a large collection of movie and TV show subtitles. It is widely used in psycholinguistics and cognitive science research. We used the PoS information in combination with the ordinal class labels of words in WorryWords to compute the distributions for each PoS.

\begin{figure}[t!]
	\centering
	    \includegraphics[width=0.47\textwidth]{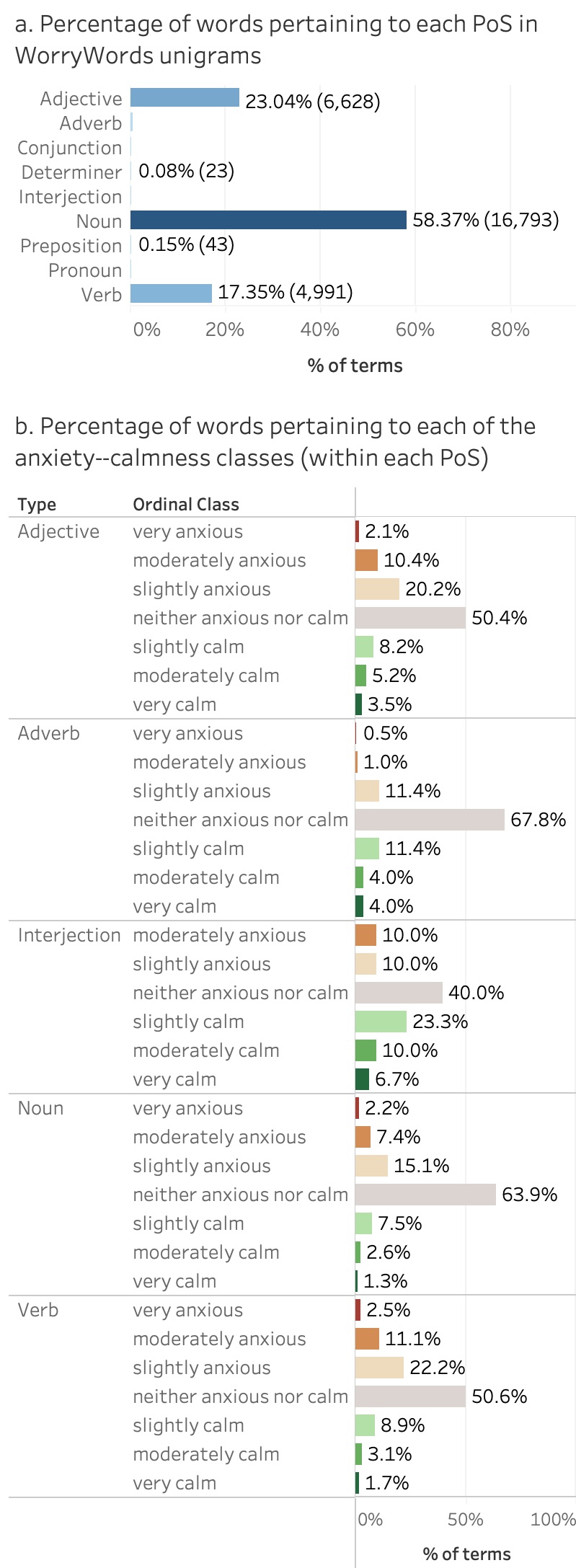}
        \caption{Unigrams - The distribution of grammatical categories  in WorryWords (unigrams) (a) and the distribution of anxiety--calmness classes within the grammatical categories of WorryWords (b).}
	    \label{fig:anx-pos}
\end{figure}

Figure \ref{fig:anx-pos} (a) shows percentage of words in WorryWords pertaining to each part of speech.
Figure \ref{fig:anx-pos} (b) shows the  percentage of words pertaining to different anxiety and calmness classes within 
each of the four main PoS classes and Interjection. Other classes were predominantly neutral.
(The numbers within each PoS sum up to 100\%.)

From (a) we see that nouns are the most frequent class in WorryWords (58\%), followed by adjectives and verbs (which are less than half as frequent). Adverbs form 1\% of the lexicon, and the rest of the classes have very small numbers. 
Interjections have a markedly highest percentage of non-neutral words (60\%), whereas conjunctions, determiners, prepositions, and pronouns are predominantly neutral. Among the more frequent PoS categories, adjectives and verbs have a markedly higher number of non-neutral words, about 50\%, which is 14 to 18\% higher than that for Nouns and Adverbs, respectively. In general, for all of these classes there is a greater percentage of anxiety associated words compared to calmness associated words; however, the trend is the opposite for interjections.\\[-7pt]

\noindent \textbf{Types of MWEs:} MWEs can be of different types such as noun compounds (noun--noun collocations), idioms/fixed expressions, particle verb constructions, etc. MWEs of each of these types are relevant to expressing anxiety and calmness. For example, \textit{walking on egg shells, bundle of nerves,} and \textit{at the edge of my rope} are idiomatic expressions conveying various levels of anxiety. Similarly, \textit{make a decision, have a breakdown, take control,} and \textit{feel pressure} are light verb constructions that convey different levels of anxiety.
Further, \textit{panic attack, stress response, worry loop, death spiral, comfort food, taste explosion, morning calm,} etc. are noun compounds conveying various levels of anxiety. Yet, we do not know the extent to which these different types of MWEs are associated with anxiety and calmness: e.g., how common is anxiety association in light verb constructions? Knowing these distributions will shed more light on how we use MWEs to express anxiety.

\begin{figure}[t]
	\centering
	    \includegraphics[width=0.49\textwidth]{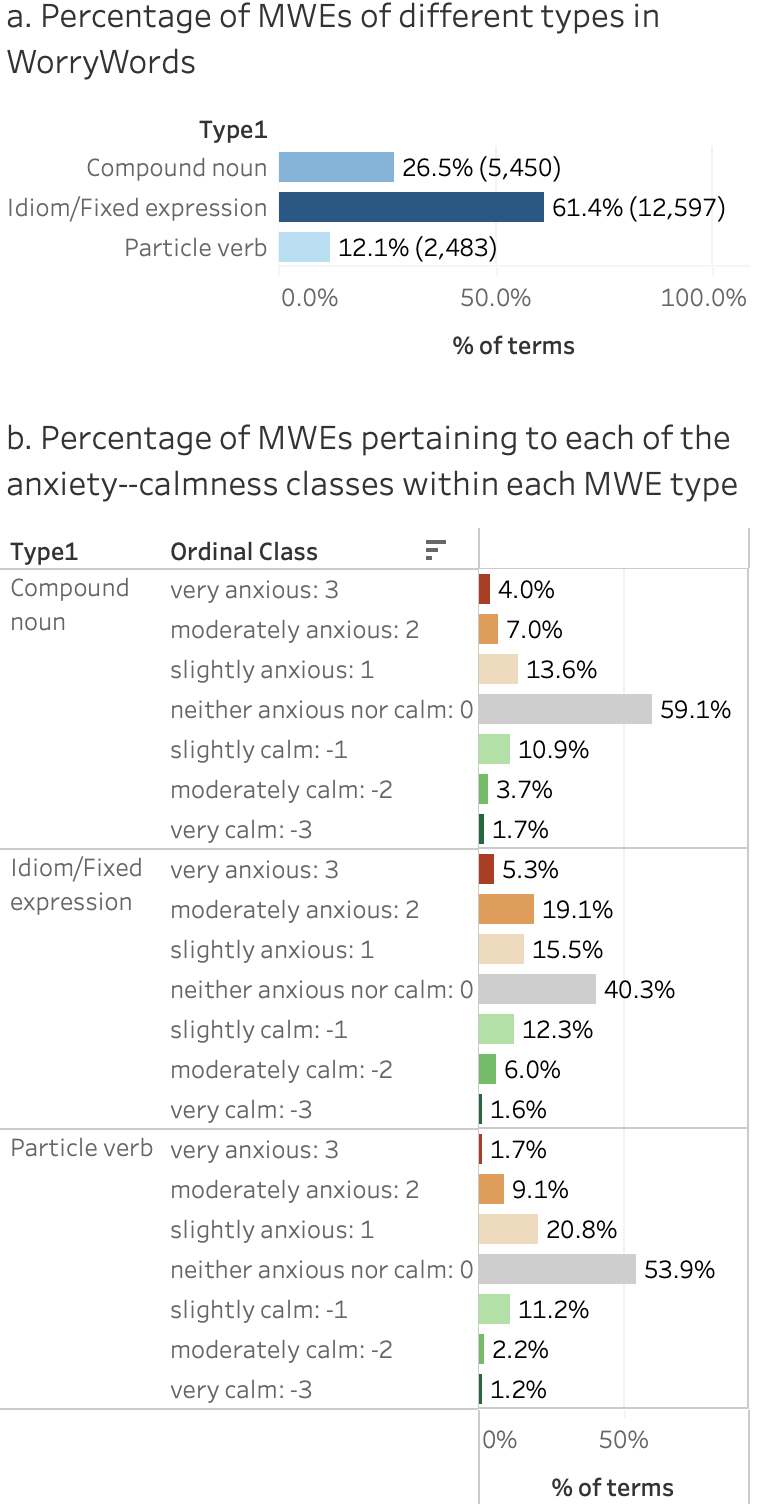}
        \caption{MWEs - The distribution of types of MWEs in WorryMWEs (a) and the distribution of anxiety--calmness classes within types of MWEs (b).}
	    \label{fig:anx-mwe-types}
\end{figure}

Each MWE entry in the MWE concreteness norms dataset \cite{muraki2023concreteness} is marked with information about its MWE type.
We make use of this to determine
 (a) percentage of different types of MWEs
--- shown in Figure \ref{fig:anx-mwe-types} (a); and (b) the  percentage of MWEs pertaining to various anxiety--calmness classes within various types of MWEs in WorryMWEs --- shown in in Figure \ref{fig:anx-mwe-types} (b). (The numbers within each type sum up to 100\%.)

\noindent \textbf{Results:} From Figure \ref{fig:anx-mwe-types} (a) we see that idioms are the most frequent class of MWEs in WorryMWEs ($\sim$61\%), followed by compound nouns and then particle verbs.
The anxiety distributions are markedly different in different types of MWES -- Figure \ref{fig:anx-mwe-types} (b). Specifically, the percentage of terms associated with some degree of anxiousness is markedly higher in idioms ($\sim$40\%) than in compound nouns ($\sim$25\%) or particle verbs ($\sim$32\%). 

Figure in the Appendix shows the distribution of different types of MWEs in bigrams, trigrams, and fourgrams, separately.
Figure in the Appendix shows the distribution of different anxiety classes in different types of MWEs in bigrams, trigrams, and fourgrams, separately.

\noindent \textbf{Types of MWEs Across Different Ngrams:} The different types of MWEs may be more or less frequent in different ngrams. Figure \ref{fig:ngrams-type-distrib} in the Appendix shows the distribution of different types of MWEs in bigrams, trigrams, and fourgrams, separately.\\
\noindent \textbf{Results:} The vast majority of noun compounds and particle verb constructions occur as bigrams; a much smaller number as trigrams, and only a handful a fourgrams.
Idioms occur most in trigrams, but also occur in good numbers  in bigrams and fourgrams (albeit about half as much as in trigrams).
We find that bigram MWEs are predominantly composed of noun compounds ($\sim$48\%), whereas trigrams and fourgrams are predominantly composed of idioms ($\sim$85\% and $\sim$99\%, respectively). 
Figure \ref{fig:ngrams-type-anx-distrib} in the Appendix shows the distribution of different anxiety classes in different types of MWE--ngram combinations (in bigrams, trigrams, and fourgrams, separately).
Notable here is that the percentage of polar idioms increases steadily (neutral class decreases) as we move from bigrams ($\sim$48\%) to trigrams ($\sim$60\%) to fourgrams ($\sim$70\%). The trend is opposite for compund nound---neutral class increases with length of ngram.

Overall, these results show that all three types of MWEs and all three types of ngrams considered are important sources of terms associated with anxiety and calmness. 

\subsection{To what extent do anxious consti- tuent words create anxious MWEs?}

A key feature of many MWEs is that they are non-compositional (to varying degrees).
This raises the question: 
to what extent is the anxiety of multiword expressions stemming from the anxiety of its constituent words; and to what extent are multiword expressions associated with anxiety/calmness when none of their constituents are associated with anxiety/calmness?

\begin{figure*}[t!]
	\centering
	    \includegraphics[width=0.9\textwidth]{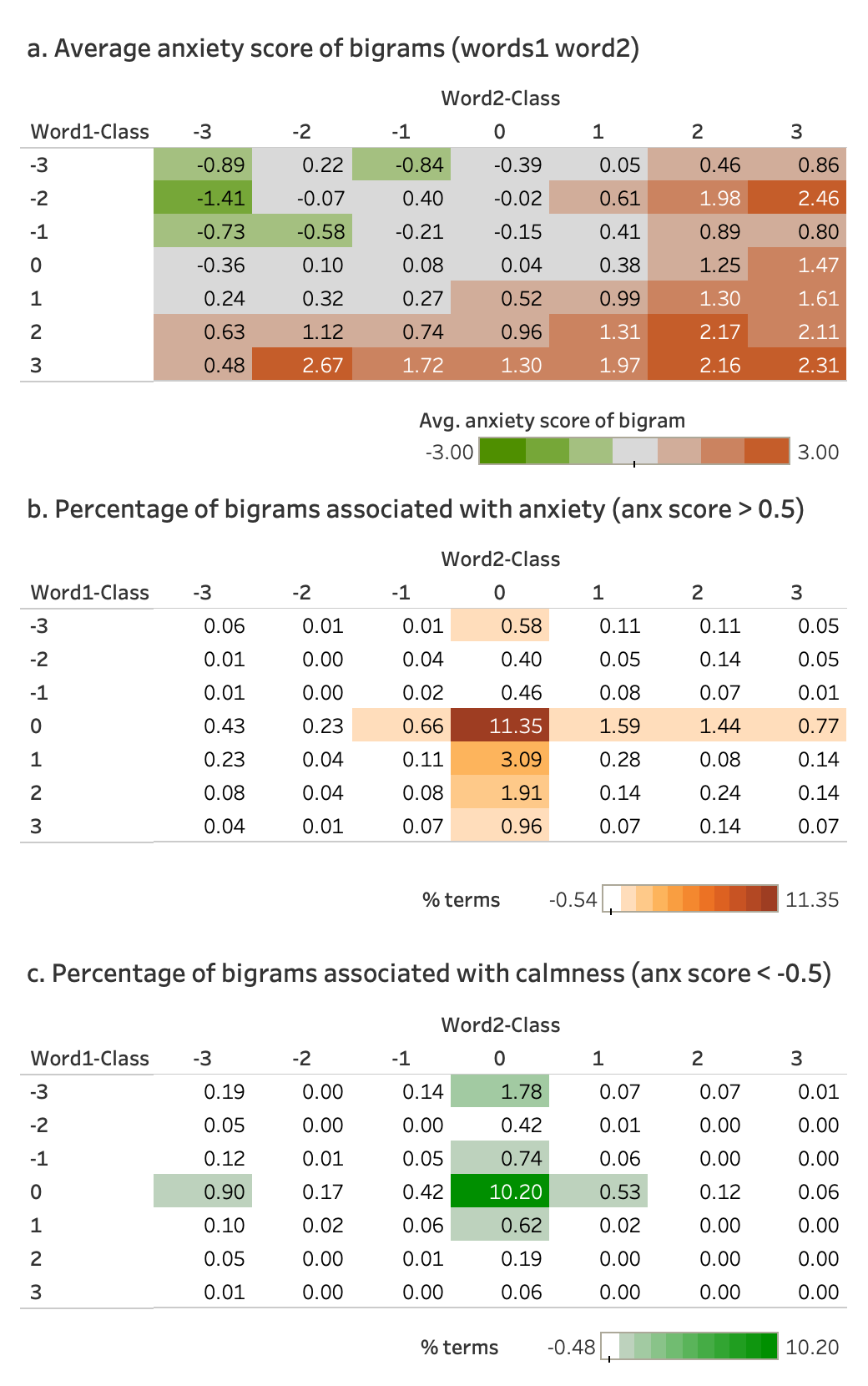}
        \caption{Measures of Anxiousness Compositionality. The percentages shown in b and c are for the 8,323 bigram MWEs considered.}
	    \label{fig:compo}
\end{figure*}

To explore this, we focused on the 8,323 bigram (two-word sequence) in WorryMWEs whose constituents are in the WorryWords lexicon. We will refer to the first word in the MWE as word1 and second as word2. We partitioned them into 49 (7*7) bins corresponding to every anxiety/calmness class combination of word1 and word2: very anxious--very anxious, very anxious--moderately anxious,..., neutral--neutral,...,vary calm--very calm.
We then determined the average anxiety scores of all the MWEs in each bin. We show the results in Figure \ref{fig:compo} (a).
For each of the 49 bins, 
we also computed the percentage of the 8,323 bigram MWEs associated with anxiety (Figure \ref{fig:compo} (b)) and the percentage of the  8,323 bigram MWEs associated with calmness (Figure \ref{fig:compo} (c)).

\noindent \textbf{Results:} Observe in (a) that as word1 and word2 bin scores increase, the average anxiety score of the bigram bins also increases. This is particularly noteworthy for word1 and word2 bin associated with anxiety (scores > 0). The trend is weaker when word1 and word2 are associated with calmness. Overall this suggests a marked degree of compositionality (more so when the constituent terms are associated with anxiety than with calmness).

However, an examination of (b) reveals that: the percentage of anxiety-associated words that have word1 and word2 associated with anxiety is relatively small ($<$0.3\%); the percentage that have exactly one of word1 and word2 associated with anxiety while the other is neutral is somewhat higher (0.75 to 1.6\%); whereas the percentage of bigrams that are associated with anxiety even though both their constituents are neutral is markedly higher (11.35\%). This shows that by far most of the bigrams associated with anxiety have neutral constituents -- indicating a substantial amount of noncompositionality.
We observe the same trend in (c) for calmness-associated bigrams. 10.2\% of the bigrams associated with calmness have constituents that are both neutral on their own.

\section{Conclusion}
We created WorryMWEs --- the first large lexicon of MWE--anxiety association scores by crowdsourcing. 
These are descriptive norms of how people use language and perceive MWE--anxiety associations.
We showed that association scores obtained from the annotations are highly reliable (split-half reliability scores between 0.81 to 0.95). 
We used WorryMWEs to 
  examine the distribution of anxiety and calmness classes in different types of words and MWEs.     We also examined the extent to which neutral and anxious/calmness terms come together to create anxious/calmness MWEs, and showed 
  that idioms are a frequent source of conveying anxiety and calmness (especially in trigram and fourgram MWEs).
Overall, we showed that MWEs are a common and important way we express anxiety and calmness.

We also examined the degree to which anxiety associations are compositional. We showed that even though some bigrams associated with anxiety have one or more constituents that are also associated with anxiety, by far, most of the MWEs associated with anxiety have neutral constituents -- indicating a substantial amount of noncompositionality.
We observed the same trend for calmness-associated MWEs. 
Finally, we made a comprehensive case for how MWE--anxiety association resources (such as WorryMWEs) are useful in wide array of disciplines, including linguistics, psychology, cognitive science, HCI, digital humanities, and NLP.
We make WorryMWEs freely available to enable a wide variety of anxiety associated research. 

This work was on English and the annotators were mostly from US, UK, India, and Canada. Future work will explore creating MWE--association lexicons for other languages. Not only will such lexicons be useful resources for those languages but also help towards a broader understanding of the role MWEs play in expressing anxiety and calmness. Work on data from many cultures will also reveal differences in how anxiety and calmness are expressed and experienced differently in different parts of the world.



\section{Limitations}
\label{sec:limitations}

WorryMWEs is the largest MWE lexicon of MWEs with annotations for anxiety; with wide coverage and a large number of annotators (hundreds of people as opposed to just a handful). 
However, people express anxiety in many creative ways, and different norms exist around anxiety expression around the world. No lexicon can cover the full range of linguistic and cultural diversity in anxiety expression. 
The lexicon is restricted to MWEs that are most commonly used in Standard American English and they capture anxiety associations as judged by American native speakers of English. 
Annotators on Mechanical Turk are not representative of the wider US population. However, obtaining annotations from a large number of annotators (as we do) makes the lexicon more resilient to individual biases and captures more diversity in beliefs.
We see this work as a first step and hopefully much more work will follow for various other groups of people and in various other languages. 
We built our lexicon using many of the principles and ideas listed in \citet{Mohammad23ethicslex}, which provides a detailed discussion of the limitations and best-practices in the creation and use of emotion lexicons.

For limitations associated with the general use of the dataset, please refer to the Ethics Statement (\S \ref{sec:ethics}).

\section{Ethics Statement}
\label{sec:ethics}

Emotions are complex and nuanced mental states. Additionally, each individual expresses  anxiety differently through language, which results in large amounts of person-to-person variation. 
We discuss below notable ethical considerations when computationally analyzing anxiety. 

Importantly, WorryMWEs should not be used as a standalone tool for detecting anxiety disorders. We hope future research, led by clinicians and medical practitioners, will assess the usefulness of WorryMWEs 
in mental health applications, in combination with various other sources of information. 

The crowd-sourced task was approved by our Institutional Research Ethics Board. 
The individual MWEs selected did not pose any risks beyond occasionally reading text on the internet. 
The annotators were free to do as many MWE annotations as they wished. The instructions included a brief description of the purpose of the task. 

Any dataset of emotion-association norms entails several ethical considerations. 
\bl{See \citet{Mohammad23ethicslex} for a discussion of good practices and ethical considerations when using emotion lexicons. See \citet{Mohammad22AER} for a broader discussion of ethical considerations relevant to automatic emotion recognition.
We discuss below notable points of discussion as well as some new and updated points especially relevant for anxiety and MWE norms.}
\begin{enumerate}
    \item \textit{Coverage:} We included a large number of English MWEs; yet, the MWEs included do not equally cover all domains, genres, and people of different locations, socio-economic strata, etc. It likely includes more of the vocabulary common in the United States with a socio-economic and educational backgrounds that allow for technology access.
  
    \item \textit{Socio-Cultural Biases:} \bl{Many multiword expressions have origins and connotations in historic racism and bigotry, e.g., {\it sold down the river, grandfathered in,} and {\it black sheep}. Many have argued that, in everyday speech, choosing alternative expressions fosters more inclusiveness. On the other other hand, use of such expressions in research can shed light on the historical and social context of racism and stereotypes permeate language.  
    }

    The annotations for anxiety capture various human biases. These biases may be systematically different for different socio-cultural groups. Our data was annotated by mostly US and Indian English speakers, but even within the US and India there are many diverse socio-cultural groups.
    Notably, crowd annotators on Amazon Mechanical Turk do not reflect populations at large. In the US for example, they tend to skew towards male, white, and younger people. However, compared to studies that involve just a handful of annotators, crowd annotations benefit from drawing on hundreds and thousands of annotators (such as this work). 

    \bl{Our dataset curation was careful to avoid words and MWEs from problematic sources. We also asked people annotate terms based on what most English speakers think (as opposed to what they themselves think). This helps to some extent, but the lexicon may still capture some historical VAD associations with certain identity groups. This can  be useful for some socio-cultural studies; but we also caution that anxiety associations with identity groups be carefully contextualized.}
    \item \textit{Perceptions (not “right” or “correct” labels):} Our goal here was to identify common perceptions of anxiety association. These are not meant to be ``correct'' or ``right'' answers, but rather what the majority of the annotators believe based on their intuitions of the English language.

    \item It is more appropriate to make claims about anxiety word and MWE usage rather than anxiety of the speakers. For example, {\it `the use of anxiety terms grew by 20\%'} rather than {\it `anxiety grew by 20\%'}. 
A marked increase in anxiety words and MWEs is likely an indication that anxiety increased, but there is no evidence that anxiety increased by 20\%.\\[-16pt]
\item Comparative analyses can be much more useful than stand-alone analyses. Often, anxiety word counts on their own are not useful. 
For example, {\it `the use of anxiety words and MWEs grew by 20\% when compared to [data from last year, data from a different person, etc.]'} is more useful than saying {\it `on average, 5 anxiety words and MWEs were used in every 100 words'}.
\item Inferences drawn from larger amounts of text are often more reliable than those drawn from small amounts of text.
 For example, {\it `the use of anxiety words and MWEs grew by 20\%'} is informative when determined from thousands, tens of thousands, or more instances. Do not draw inferences about a single sentence or utterance from the anxiety associations of its constituent words.     
\end{enumerate}
\noindent We recommend careful reflection of ethical considerations relevant for the specific context of deployment when using WorryMWEs.

The use of the \newcite{muraki2023concreteness} MWE dataset was in accordance with its terms of use. They have made the resource freely available for research.


\section{Bibliographical References}
\label{sec:reference}
\bibliographystyle{lrec2026-natbib}
\bibliography{custom}

@string{ACL:2018:1 = {Proceedings of the 56th Annual Meeting of the Association for Computational Linguistics (Volume 1: Long Papers)}}

@string{LREC:2018:1 = {Proceedings of the Eleventh International Conference on Language Resources and Evaluation ({LREC} 2018)}}

@string{CLPSYCH:2024:1 = {Proceedings of the 9th Workshop on Computational Linguistics and Clinical Psychology (CLPsych 2024)}}

@string{EMNLP:2019:62 = {Proceedings of the Tenth International Workshop on Health Text Mining and Information Analysis (LOUHI 2019)}}

@string{EMNLP:2024:main = {Proceedings of the 2024 Conference on Empirical Methods in Natural Language Processing}}

@string{acl = {Association for Computational Linguistics}}

@string{anth = {https://aclanthology.org/}}

@misc{WorryLex,
  author       = {Saif M. Mohammad},
  title        = {From Composure to Catastrophe: 
Norms of Calmness--Anxiety Associations for 64,000 English Words and Multiword Expressions},
  year         = {2026},
  howpublished = {PsyArXiv},
  doi          = {10.31234/osf.io/qc6gr_v1}
}

@article{constant-etal-2017-survey,title = "{S}urvey: Multiword Expression Processing: A {S}urvey",author = {Constant, Mathieu and Eryiǧit, G{\"u}l{\c{s}}en and Monti, Johanna and van der Plas, Lonneke and Ramisch, Carlos and Rosner, Michael and Todirascu, Amalia},journal = "Computational Linguistics",volume = "43",number = "4",month = dec,year = "2017",address = "Cambridge, MA",publisher = "MIT Press",url = anth # {J17-4005/},doi = "10.1162/COLI_a_00302",pages = "837--892"}

@inproceedings{mohammad-2018-obtaining,title = "Obtaining Reliable Human Ratings of Valence, Arousal, and Dominance for 20,000 {E}nglish Words",author = "Mohammad, Saif",editor = "Gurevych, Iryna and Miyao, Yusuke",booktitle = ACL:2018:1,month = jul,year = "2018",address = "Melbourne, Australia",publisher = acl,url = anth # {P18-1017/},doi = "10.18653/v1/P18-1017",pages = "174--184"}

@inproceedings{jochim-etal-2018-slide,title = "{SLIDE} - a Sentiment Lexicon of Common Idioms",author = "Jochim, Charles and Bonin, Francesca and Bar-Haim, Roy and Slonim, Noam",editor = "Calzolari, Nicoletta and Choukri, Khalid and Cieri, Christopher and Declerck, Thierry and Goggi, Sara and Hasida, Koiti and Isahara, Hitoshi and Maegaard, Bente and Mariani, Joseph and Mazo, H{\'e}l{\`e}ne and Moreno, Asuncion and Odijk, Jan and Piperidis, Stelios and Tokunaga, Takenobu",booktitle = LREC:2018:1,month = may,year = "2018",address = "Miyazaki, Japan",publisher = "European Language Resources Association (ELRA)",url = anth # {L18-1379/}}

@inproceedings{mitrovic-etal-2024-comparing,title = "Comparing panic and anxiety on a dataset collected from social media",author = "Mitrovi{\'c}, Sandra and Lithgow-Serrano, Oscar William and Schillaci, Carlo",editor = "Yates, Andrew and Desmet, Bart and Prud{'}hommeaux, Emily and Zirikly, Ayah and Bedrick, Steven and MacAvaney, Sean and Bar, Kfir and Ireland, Molly and Ophir, Yaakov",booktitle = CLPSYCH:2024:1,month = mar,year = "2024",address = "St. Julians, Malta",publisher = acl,url = anth # {2024.clpsych-1.12/},pages = "153--165"}

@inproceedings{turcan-mckeown-2019-dreaddit,title = "{D}readdit: A {R}eddit Dataset for Stress Analysis in Social Media",author = "Turcan, Elsbeth and McKeown, Kathy",editor = "Holderness, Eben and Jimeno Yepes, Antonio and Lavelli, Alberto and Minard, Anne-Lyse and Pustejovsky, James and Rinaldi, Fabio",booktitle = EMNLP:2019:62,month = nov,year = "2019",address = "Hong Kong",publisher = acl,url = anth # {D19-6213/},doi = "10.18653/v1/D19-6213",pages = "97--107"}

@inproceedings{mohammad-2024-worrywords,title = "{W}orry{W}ords: Norms of Anxiety Association for over 44k {E}nglish Words",author = "Mohammad, Saif M.",editor = "Al-Onaizan, Yaser and Bansal, Mohit and Chen, Yun-Nung",booktitle = EMNLP:2024:main,month = nov,year = "2024",address = "Miami, Florida, USA",publisher = acl,url = anth # {2024.emnlp-main.910/},doi = "10.18653/v1/2024.emnlp-main.910",pages = "16261--16278"}

@article{yeste2026schwartz,
  title={Do Schwartz Higher-Order Values Help Sentence-Level Human Value Detection? When Hard Gating Hurts},
  author={Yeste, V{\'\i}ctor and Rosso, Paolo},
  journal={arXiv preprint arXiv:2602.00913},
  year={2026}
}

@misc{mazhar2025memes,
      title={Figurative-cum-Commonsense Knowledge Infusion for Multimodal Mental Health Meme Classification}, 
      author={Abdullah Mazhar and Zuhair hasan shaik and Aseem Srivastava and Polly Ruhnke and Lavanya Vaddavalli and Sri Keshav Katragadda and Shweta Yadav and Md Shad Akhtar},
      year={2025},
      eprint={2501.15321},
      archivePrefix={arXiv},
      primaryClass={cs.CL},
      url={https://arxiv.org/abs/2501.15321}, 
}

@inproceedings{mohammad2025breaking,
  title={Breaking Bad: Norms for Valence, Arousal, and Dominance for over 10k English Multiword Expressions},
  author={Mohammad, Saif},
  booktitle={Proceedings of the 14th International Joint Conference on Natural Language Processing and the 4th Conference of the Asia-Pacific Chapter of the Association for Computational Linguistics},
  pages={1964--1988},
  year={2025}
}

@misc{mohammad2026temporalanxiety,
      title={When are We Worried? Temporal Trends of Anxiety and What They Reveal about Us}, 
      author={Saif M. Mohammad},
      year={2026},
      eprint={2602.10400},
      archivePrefix={arXiv},
      primaryClass={cs.CL},
      url={https://arxiv.org/abs/2602.10400}, 
}

@article{ramisch2015multiword,
  title={Multiword expressions acquisition},
  author={Ramisch, Carlos},
  journal={A Generic and Open Framework. Cham: Springer International Publishing},
  year={2015},
  publisher={Springer}
}

@article{citron2019idiomatic,
  title={Idiomatic expressions evoke stronger emotional responses in the brain than literal sentences},
  author={Citron, Francesca MM and Cacciari, Cristina and Funcke, Jakob M and Hsu, Chun-Ting and Jacobs, Arthur M},
  journal={Neuropsychologia},
  volume={131},
  pages={233--248},
  year={2019},
  publisher={Elsevier}
}

@book{smolka2020role,
  title={The role of constituents in multiword expressions: An interdisciplinary, cross-lingual perspective (Volume 4)},
  author={Smolka, Eva and Schulte im Walde, Sabine},
  year={2020},
  publisher={Language Science Press}
}

@article{kacinik2014sticking,
  title={Sticking your neck out and burying the hatchet: what idioms reveal about embodied simulation},
  author={Kacinik, Natalie A},
  journal={Frontiers in human neuroscience},
  volume={8},
  pages={689},
  year={2014},
  publisher={Frontiers Media SA}
}

@article{allawama2025idioms,
  title={Idioms as Gateways to Emotional Expressions of Sadness and Joy in French},
  author={Allawama, Ashraf and Zibin, Aseel and Altakhaineh, Abdel Rahman and others},
  journal={Journal of Intercultural Communication},
  volume={25},
  number={1},
  pages={83--97},
  year={2025},
  publisher={International Collaboration for Research \& Publications}
}

@article{nichter2010idioms,
  title={Idioms of distress revisited},
  author={Nichter, Mark},
  journal={Culture, Medicine, and Psychiatry},
  volume={34},
  number={2},
  pages={401--416},
  year={2010},
  publisher={Springer}
}

@inproceedings{lee2017figurative,
  title={Figurative language in emotion expressions},
  author={Lee, Sophia Yat Mei},
  booktitle={Workshop on Chinese Lexical Semantics},
  pages={408--419},
  year={2017},
  organization={Springer}
}

@article{cole2010role,
  title={The role of language in the development of emotion regulation},
  author={Cole, Pamela M and Armstrong, Laura Marie and Pemberton, Caroline K},
  journal={Child development at the intersection of emotion and cognition},
  year={2010},
  publisher={American Psychological Association}
}

@book{cap2016language,
  title={The language of fear: Communicating threat in public discourse},
  author={Cap, Piotr},
  year={2016},
  publisher={Springer}
}

@article{ibrahim2015idioms,
  title={Idioms-proverbs lexicon for modern standard Arabic and colloquial sentiment analysis},
  author={Ibrahim, Hossam S and Abdou, Sherif M and Gheith, Mervat},
  journal={arXiv preprint arXiv:1506.01906},
  year={2015}
}

@article{tong2024metaphor,
  title={Metaphor understanding challenge dataset for LLMs},
  author={Tong, Xiaoyu and Choenni, Rochelle and Lewis, Martha and Shutova, Ekaterina},
  journal={arXiv preprint arXiv:2403.11810},
  year={2024}
}

@article{heinrichs2006cultural,
  title={Cultural differences in perceived social norms and social anxiety},
  author={Heinrichs, Nina and Rapee, Ronald M and Alden, Lynn A and B{\"o}gels, Susan and Hofmann, Stefan G and Oh, Kyung Ja and Sakano, Yuji},
  journal={Behaviour research and therapy},
  volume={44},
  number={8},
  pages={1187--1197},
  year={2006},
  publisher={Elsevier}
}

@inproceedings{villavicencio2004lexical,
  title={Lexical encoding of MWEs},
  author={Villavicencio, Aline and Copestake, Ann and Waldron, Benjamin and Lambeau, Fabre},
  booktitle={Proceedings of the Workshop on Multiword Expressions: Integrating Processing},
  pages={80--87},
  year={2004}
}

@article{jackendoff1997twistin,
  title={Twistin'the night away},
  author={Jackendoff, Ray},
  journal={Language},
  pages={534--559},
  year={1997},
  publisher={JSTOR}
}

@article{zgusta1967multiword,
  title={Multiword lexical units},
  author={Zgusta, Ladislav},
  journal={Word},
  volume={23},
  number={1-3},
  pages={578--587},
  year={1967},
  publisher={Taylor \& Francis}
}

@inproceedings{sag2002multiword,
  title={Multiword expressions: A pain in the neck for NLP},
  author={Sag, Ivan A and Baldwin, Timothy and Bond, Francis and Copestake, Ann and Flickinger, Dan},
  booktitle={International conference on intelligent text processing and computational linguistics},
  pages={1--15},
  year={2002},
  organization={Springer}
}

@incollection{takahashi2024comprehensive,
  title={A comprehensive Japanese MWE lexicon: JMWEL},
  author={Takahashi, Masahito and Tanabe, Toshifumi and Halpern, Jack and Shudo, Kosho},
  booktitle={Recent Advances in Multiword Units in Machine Translation and Translation Technology},
  pages={218--242},
  year={2024},
  publisher={John Benjamins Publishing Company}
}

@article {PMID:35867207,
	Title = {Concreteness ratings for 62,000 English multiword expressions},
	Author = {Muraki, Emiko J and Abdalla, Summer and Brysbaert, Marc and Pexman, Penny M},
	DOI = {10.3758/s13428-022-01912-6},
	Number = {5},
	Volume = {55},
	Month = {August},
	Year = {2023},
	Journal = {Behavior research methods},
	ISSN = {1554-351X},
	Pages = {2522—2531},
	URL = {https://doi.org/10.3758/s13428-022-01912-6}
}

@article{brysbaert2009moving,
  title={Moving beyond Ku{\v{c}}era and Francis: A critical evaluation of current word frequency norms and the introduction of a new and improved word frequency measure for American English},
  author={Brysbaert, Marc and New, Boris},
  journal={Behavior research methods},
  volume={41},
  number={4},
  pages={977--990},
  year={2009},
  publisher={Springer}
}

@inproceedings{worrywords-emnlp2024,
  title={WorryWords: Norms of Anxiety Association for 44,450 English Words},
  author={Mohammad, Saif M.},
    booktitle={Proceedings of The Annual Conference of the Empirical Methods on Natural Language Processing (EMNLP 2024, main)},
    year={2024},
    address={Miami, FL}
}

@article{brysbaert2012adding,
  title={Adding part-of-speech information to the SUBTLEX-US word frequencies},
  author={Brysbaert, Marc and New, Boris and Keuleers, Emmanuel},
  journal={Behavior research methods},
  volume={44},
  pages={991--997},
  year={2012},
  publisher={Springer}
}

@article{muraki2023concreteness,
  title={Concreteness ratings for 62,000 English multiword expressions},
  author={Muraki, Emiko J and Abdalla, Summer and Brysbaert, Marc and Pexman, Penny M},
  journal={Behavior research methods},
  volume={55},
  number={5},
  pages={2522--2531},
  year={2023},
  publisher={Springer}
}

@article{Mohammad22AER,
  title={Ethics Sheet for Automatic Emotion Recognition and Sentiment Analysis},
  author={Mohammad, Saif M.},
  journal={Computational Linguistics},
     volume = {48},
    number = {2},
    pages = {239-278},
   month     = {June},
  year={2022}
}

@inproceedings{Mohammad23ethicslex,
      title={Best Practices in the Creation and Use of Emotion Lexicons},
      author={Mohammad, Saif M.},
      year={2023},
      address = {Dubrovnik, Croatia},
      publisher = "Association for Computational Linguistics",
      booktitle = "Proceedings of the 17th Conference of the European Chapter of the Association for Computational Linguistics"
}

@article{siddaway2018reconceptualizing,
  title={Reconceptualizing Anxiety as a Continuum That Ranges From High Calmness to High Anxiety: The Joint Importance of Reducing Distress and Increasing Well-Being.},
  author={Siddaway, Andy P and Taylor, Peter J and Wood, Alex M},
  journal={Journal of personality and social psychology},
  volume={114},
  number={2},
  pages={e1},
  year={2018},
  publisher={American Psychological Association}
}

@article{kim2005psychobiology,
  title={The psychobiology of anxiety},
  author={Kim, Jean and Gorman, Jack},
  journal={Clinical Neuroscience Research},
  volume={4},
  number={5-6},
  pages={335--347},
  year={2005},
  publisher={Elsevier}
}

@article{bateson2011anxiety,
  title={Anxiety: an evolutionary approach},
  author={Bateson, Melissa and Brilot, Ben and Nettle, Daniel},
  journal={The Canadian Journal of Psychiatry},
  volume={56},
  number={12},
  pages={707--715},
  year={2011},
  publisher={SAGE Publications Sage CA: Los Angeles, CA}
}

@article{MohammadT13,
author = {Mohammad, Saif M. and Turney, Peter D.},
title = {Crowdsourcing a Word--Emotion Association Lexicon},
journal = {Computational Intelligence},
publisher = {Wiley Blackwell Publishing Ltd.},
volume = {29},
number = {3},
pages = {436--465},
year = {2013}
}

@article{warriner2013norms,
  title={Norms of valence, arousal, and dominance for 13,915 {E}nglish lemmas},
  author={Warriner, Amy Beth and Kuperman, Victor and Brysbaert, Marc},
  journal={Behavior Research Methods},
  volume={45},
  number={4},
  pages={1191--1207},
  year={2013},
  publisher={Springer}
}

@article{weir2005quantifying,
  title={Quantifying test-retest reliability using the intraclass correlation coefficient and the SEM},
  author={Weir, Joseph P},
  journal={The Journal of Strength \& Conditioning Research},
  volume={19},
  number={1},
  pages={231--240},
  year={2005},
  publisher={LWW}
}

@inproceedings{MohammadT10,
  author = {Mohammad, Saif M. and Turney, Peter D.},
  title = {Emotions Evoked by Common Words and Phrases: Using {Mechanical Turk} to Create an Emotion Lexicon},
  booktitle = {Proceedings of the NAACL-HLT Workshop on Computational Approaches to Analysis and Generation of Emotion in Text},
  address={LA, California},
  year = {2010}}

@article{moors2013norms,
  title={Norms of valence, arousal, dominance, and age of acquisition for 4,300 Dutch words},
  author={Moors, Agnes and De Houwer, Jan and Hermans, Dirk and Wanmaker, Sabine and Van Schie, Kevin and Van Harmelen, Anne-Laura and De Schryver, Maarten and De Winne, Jeffrey and Brysbaert, Marc},
  journal={Behavior research methods},
  volume={45},
  number={1},
  pages={169--177},
  year={2013},
  publisher={Springer}
}

@article{vo2009berlin,
  title={The Berlin affective word list reloaded (BAWL-R)},
  author={V{\~o}, Melissa LH and Conrad, Markus and Kuchinke, Lars and Urton, Karolina and Hofmann, Markus J and Jacobs, Arthur M},
  journal={Behavior research methods},
  volume={41},
  number={2},
  pages={534--538},
  year={2009},
  publisher={Springer}
}

@INPROCEEDINGS{rastogi2022stress,
  author={Rastogi, Aryan and Qian, Liu and Cambria, Erik},
  booktitle={2022 IEEE World Conference of Computational Intelligence (WCCI).},
  title={Stress Detection from Social Media Articles: New Dataset Benchmark and Analytical Study},
  year={2022},
  volume={},
  number={},
  pages={},
  doi={},
  ISSN={},
  month={}
}

 \appendix

\section*{Appendix}
Supplementary figures. 

 \begin{figure*}[t]
	     \centering
	     \includegraphics[width=\textwidth]{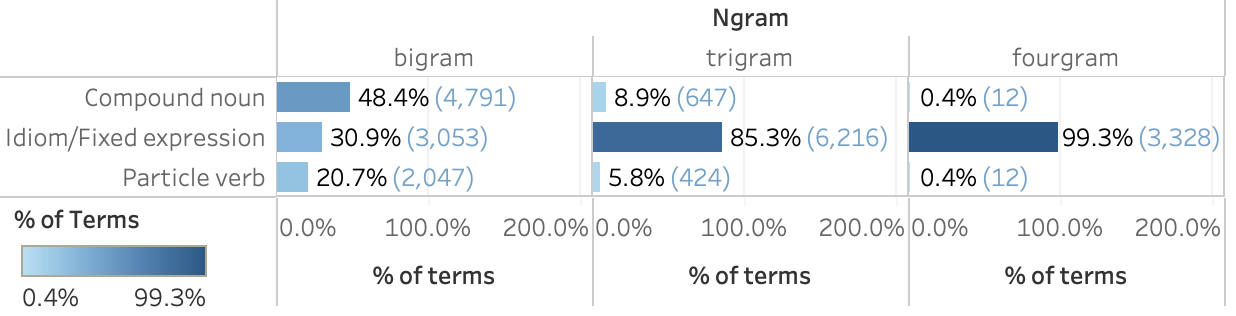}
      \caption{Distribution of  Types of MWEs across different ngrams in WorryMWEs: Percentage of terms of a given ngram (bigram, trigram, or fourgram) associated with each MWE type. (The total number is shown in parenthesis.)}
	     \label{fig:ngrams-type-distrib}
	 \end{figure*}

 \begin{figure*}[t]
	     \centering
	     \includegraphics[width=\textwidth]{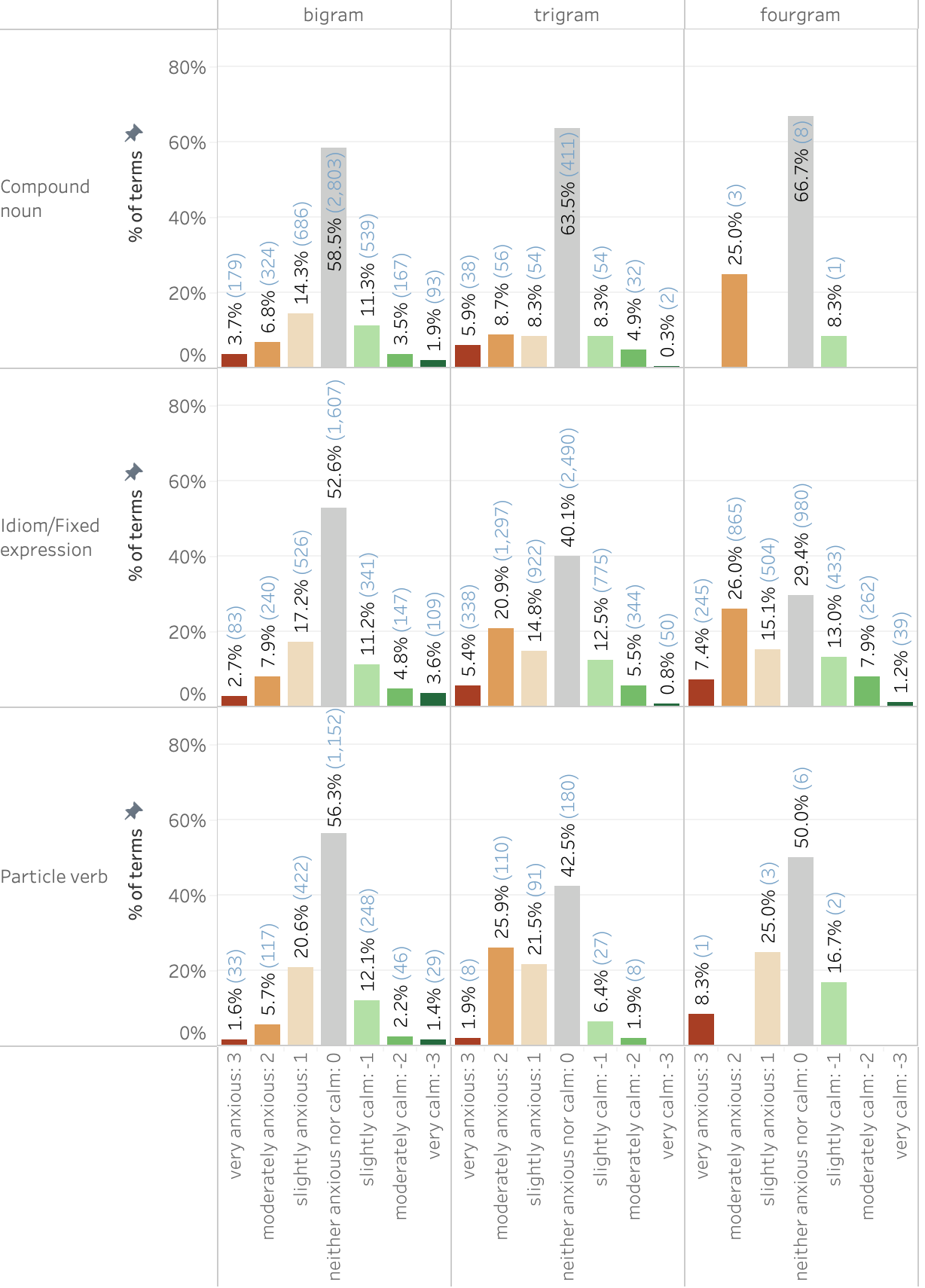}
      \caption{Distribution of anxiety and calmness classes in each of the MWE types and ngram pairs: percentage of terms associated with each class. (The total number is shown in parenthesis.)}
	     \label{fig:ngrams-type-anx-distrib}
	 \end{figure*}

\end{document}